\documentclass[a4paper]{article}

\usepackage{INTERSPEECH2021}

\title{Improving Distinction between ASR Errors and Speech Disfluencies \break with Feature Space Interpolation}
\name{Seongmin Park, Dongchan Shin, Sangyoun Paik, Subong Choi, Alena Kazakova, Jihwa Lee}
\address{ActionPower, Seoul, Republic of Korea}
\email{\{seongmin.park, dongchan.shin, sangyoun.paik, subong.choi, alena.kazakova, jihwa.lee\}@actionpower.kr}

\begin{document}

\maketitle
\begin{abstract}
    Fine-tuning pretrained language models (LMs) is a popular approach to automatic speech recognition (ASR) error detection during post-processing. While error detection systems often take advantage of statistical language archetypes captured by LMs, at times the pretrained knowledge can hinder error detection performance. For instance, presence of speech disfluencies might confuse the post-processing system into tagging disfluent but accurate transcriptions as ASR errors. Such confusion occurs because both error detection and disfluency detection tasks attempt to identify tokens at statistically unlikely positions. This paper proposes a scheme to improve existing LM-based ASR error detection systems, both in terms of detection scores and resilience to such distracting auxiliary tasks. Our approach adopts the popular \textit{mixup} method in text feature space and can be utilized with any black-box ASR output. To demonstrate the effectiveness of our method, we conduct post-processing experiments with both traditional and end-to-end ASR systems (both for English and Korean languages) with 5 different speech corpora. We find that our method improves both ASR error detection $F_1$ scores and reduces the number of correctly transcribed disfluencies wrongly detected as ASR errors. Finally, we suggest methods to utilize resulting LMs directly in semi-supervised ASR training. 
\end{abstract}

\section{Introduction}

Post-processing is the final auditor of an automatic speech recognition (ASR) system before its output is presented to consumers or downstream tasks. Two notable tasks often entrusted to the post-processing step are ASR error detection \cite{feld_mobile_2012, ogrodniczuk2020proceedings, mani_asr_2020} and speech disfluency detection \cite{shriberg_preliminaries_1994, salesky_towards_2018, wang_semi-supervised_2018}. 

A recent trend in post-processing is to fine-tune pretrained transformer-based language models (LMs)  such as BERT \cite{devlin_bert_2019} or GPT-2 \cite{radford2019language} for designated tasks. Utilizing language patterns captured by huge neural networks during their unsupervised pretraining yields considerable advantages in downstream language tasks \cite{weng_joint_2020, bach_noisy_2019, hrinchuk_correction_2020}. When using pretrained LMs, both ASR error correction and speech disfluency detection tasks employ a similar setup, where the LM is further fine-tuned with task-specific data and label, often in a sequence-labeling scheme \cite{lee_auxiliary_2020, Tanaka2018NeuralEC}.  Disfluency in speech, by definition, consists of sequence of tokens unlikely to appear together. ASR errors are statistically likely to introduce tokens heavily out of context. Both descriptions resemble the pretraining objective of LMs, which is to learn a language\textquotesingle s statistical landscape. 

From such observation, a natural question follows: in a situation where objectives of two post-processing tasks are mutually exclusive (but both attune to the LM pretraining objective), will utilizing a pretrained LM benefit a single task? Such concerns arise especially in production settings, where customer orations tend to contain significantly more disfluencies than in audio found in academic benchmarks. This situation is illustrated in Figure 1. In the figure, ASR output of disfluent speech (\textit{ASR output}) is compared with its reference transcript (\textit{Ground truth}). Words \textit{like} and \textit{um} in the reference are annotated as disfluent. The post-processing system must only detect ASR errors (red and blue), while refraining from detecting merely disfluent speech (green) as ASR errors.

\begin{figure}[t]
  \centering
  \includegraphics[width=\linewidth]{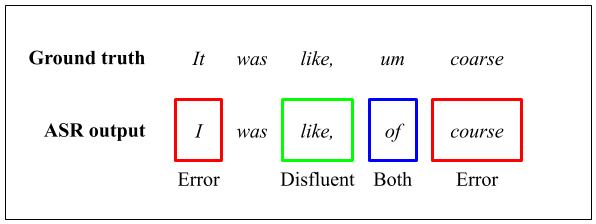}
  \caption{Demonstration of speech disfluency and ASR error.}
  \label{fig:example}
\end{figure}
This paper proposes a regularization scheme that improves ASR error detection performance by weakening an LM’s propensity to tag accurate transcriptions of disfluent speech as ASR errors. Our method is applied during the LM fine-tuning stage. Resulting LMs have been applied as drop-in replacements in existing ASR error correction systems in production.

Furthermore, LM artifacts from our method can be utilized beyond post-prcessing. Recent advancements in semi-supervised ASR training take advantage of deep LMs to create pseudo-labels for unlabeled audio \cite{park2020improved, hsu2020semi}. To ensure the quality of synthetic labels, \cite{park2020improved} uses an LM during beam-search decoding; \cite{hsu2020semi} filters created labels by length-normalized language model scores. In both cases, an LM robust to speech disfluencies is essential in improving transcription accuracy. 

\section{Related work}

\subsection{ASR post-processing}
Post-processing is a task that complements ASR systems by analyzing their output. Compared to tasks involved in directly manipulating components within the ASR system, the black-box approach of ASR post-processing yields several advantages. It is often intractable for the LM inside an ASR system to account for and rank all hypothesis presented by the acoustic model \cite{ogrodniczuk_proceedings_2020}. A post-processing system is free from such constraints within an ASR system \cite{feld_mobile_2012}. ASR post-processing also helps improve performance in downstream tasks \cite{wang_data_2020} as well as in domain adaptation flexibility \cite{anantaram_adapting_2017}. 

\subsubsection{ASR error detection}
One common approach to ASR error detection is through machine translation. Machine translation approaches in error detection commonly interpret error detection as a sequence-to-sequence task, and construct a system to translate erroneous transcriptions to correct sentences \cite{mani_asr_2020, hrinchuk_correction_2020, mani_towards_2020}. \cite{mani_asr_2020} suggests a method applicable to off-the-shelf ASR systems. \cite{hrinchuk_correction_2020} and \cite{mani_towards_2020} translates acoustic model outputs directly to grammatical text.  

In some domains, however, necessity of huge parallel training data can render translation-based error detection infeasible \cite{popel_training_2018}. Another popular approach models ASR error detection as a sequence labeling problem. This approach requires much less training samples compared to sequence-to-sequence systems. Both \cite{weng_joint_2020} and \cite{tanaka_neural_nodate} use a large, pretrained deep LM to gauge the likeliness of each output token from an ASR system. 

\subsubsection{Speech disfluency detection}
The aim of a disfluency detection task is to separate well-formed sentences from its unnatural counterparts. Disfluencies in speech include deviant language, such as hesitation, repetition, correction, and false starts \cite{salesky_towards_2018}. Speech disfluencies display regularities that can be detected with statistical systems \cite{shriberg_preliminaries_1994}.  While various suggestions such as semi-supervised \cite{wang_semi-supervised_2018} and self-supervised \cite{wang_multi-task_2020, wang_combining_2020} approaches were proposed, supervised training on pretrained LM remains state-of-the-art \cite{bach_noisy_2019, lee_auxiliary_2020}. Both \cite{bach_noisy_2019} and \cite{lee_auxiliary_2020} fine-tunes deep LMs for sequence tagging. 

\subsection{Data interpolation for regularization}
Interpolating existing data for data augmentation is a popular method for regularization. The \textit{mixup} method suggested in \cite{zhang_mixup_2018} has been particularly effective, especially in the computer vision domain \cite{he_bag_2019, berthelot2019mixmatch, bochkovskiy_yolov4_2020}. Given dataset $D = \{(x_0, y_0), (x_1, y_1) \ldots, (x_n, y_n)\}$ consisting of \textit{(data, label)} pairs, \textit{mixup} “mixes” two random training samples $(x_i, y_i)$ and $(x_j, y_j)$ to generate a new training sample $(x_m, y_m)$:
\begin{equation}
  x_m = x_i * \lambda + x_j * (1-\lambda)
  \label{eq1}
\end{equation}
\begin{equation}
    y_m = y_i * \lambda + y_j * (1-\lambda)
    \label{eq2}
\end{equation}
where
\begin{equation*}
     \lambda \sim Beta(\alpha, \alpha)
\end{equation*}

$\alpha$ is a hyperparameter for the \textit{mixup} operation that characterizes the Beta distribution to randomly sample $\lambda$ from.

\cite{verma_manifold_2019} suggests \textit{manifold mixup} to utilize \textit{mixup} in the feature space after the data has been passed through a vectorizing layer, such as a deep neural network (DNN). Whereas \textit{mixup} explicitly generates new data by interpolating existing samples in the data space, \textit{manifold mixup} adds a regularization layer to an existing architecture that mixes data in its vectorized form. In essence, \textit{mixup} and its variants discourage a model from making extreme decisions. A hidden space visualization in \cite{verma_manifold_2019} shows that correct targets and non-correct targets are better separated within the \textit{mixup} model because it learns to handle intermediate values instead of making wild guesses in face of uncertainty.   

Because text is both ordered and discrete, \textit{mixup} with text data is best applied at feature space. Magnitudes of each token id in a tokenized text sequence does not convey any relative meaning; whereas in an image, varying the degrees of continuous pixel values does gradually manipulate its properties. 

While \textit{mixup} is yet to be widely studied in speech-to-text or natural language processing domains, several studies have found success in applying the method to text data. \cite{guo_sequence-level_2020} constructs a new token sequence by randomly selecting a token from two different sequences at each index. \cite{zhang_seqmix_2020} suggests applying \textit{mixup} in feature space, after an intermediary layer of a pretrained LM. New input data is then generated by reversing the synthetic feature to find the most similar token in the vocabulary. While these two works involve interpolating text inputs, our method differs significantly in that we do not directly generate augmented training samples; instead, we utilize \textit{mixup} as a regularizing layer during the training process. Our method also does not require reversing word-embeddings or discriminative filtering using GPT-2 introduced in \cite{zhang_seqmix_2020}.

\cite{kong_calibrated_2020}, a work most similar to ours, applies \textit{manifold mixup} on text data to calibrate a model to work best on out-of-distribution data. However, the study focuses on sentence-level classification, which requires a substantially different feature extraction strategy from the work in this paper. Our paper suggests a token-level \textit{mixup} method, and aims to gauge the impact of single task (ASR error detection) regularization on an auxiliary task (speech disfluency detection). 

\section{Methodology}

\subsection{BERT for sequence tagging}
To utilize BERT for a sequence labeling problem, a token classification head (TCH) must be added after BERT’s final layer. There is no restriction on what constitutes a TCH. In \cite{lee_auxiliary_2020}, for example, a combination of conditional random field and a fully connected network was attached to BERT’s output as a TCH. In our study, a simple fully-connected layer with the same number of inputs and outputs as the original input token sequence suffices as a TCH because our aim is to measure the effectiveness of regularization that takes place between BERT and the TCH. We modify a reference BERT implementation \cite{wolf_huggingfaces_2019} with a TCH.

A single data sample during the supervised training stage consists of input token sequence $\vec{x} = {x_0, x_1, \ldots , x_n}$, and a sequence of labels for each token, $\vec{y} = {y_0, y_1 \ldots, y_n}$. Each corresponding label has a value of 1 if the token contains a character with erroneous ASR output, and 0 if the token contains no wrong character. All input sequences are tokenized into subword tokens \cite{sennrich2016neural} before entering BERT. For the rest of this paper, boldfaced letters denote vectors.

\subsection{Feature space interpolation}
Our proposed method adds a regularizing layer between BERT’s output and the token classification head. BERT produces a hidden state sequence $\vec{h}$, with length equal to that of input token sequence $\vec{x}$: 

\begin{equation}
  \vec{h} = {h_0, h_1, \ldots , h_n} = BERT(\vec{x})
  \label{eq3}
\end{equation}

From $\vec{h}$ we obtain $\vec{h_{shuffled}}$, a randomly permuted version of $\vec{h}$. Each corresponding label in $\vec{y}$ is also rearranged accordingly, creating a new label vector $\vec{y_{shuffled}}$. The same index-wise shuffle applied to $\vec{h}$ applies to $\vec{y}$, such that each label $y_i$ originally assigned to an input token $x_i$ follows the new index for that value during the shuffling process: 
\begin{equation*}
  \vec{h_{shuffled}} = {h_{s0}, h_{s1}, \ldots , h_{sn}}
\end{equation*}
\begin{equation*}
  \vec{y_{shuffled}} = {y_{s0}, y_{s1} \ldots, y_{sn}}
\end{equation*}
where
\begin{equation*}
  s0 \ldots sn = Random Permutation(0...n)
\end{equation*}
Next, we apply the \textit{mixup} process element-wise to $\vec{h}$ and $\vec{h_{shuffled}}$, according to equation (1) and (2): 
\begin{equation}
  \vec{h_{new}} = (\vec{h} \odot \lambda) + (\vec{h_{shuffled}} \odot (1 - \lambda))
  \label{eq4}
\end{equation}
Similarly, we also mix the original label vector $\vec{y}$ with its permuted counterpart $\vec{y_{shuffled}}$: 
\begin{equation}
  \vec{y_{new}} = (\vec{y} \odot \lambda) + (\vec{y_{shuffled}} \odot (1 - \lambda))
  \label{eq5}
\end{equation}
Where $\lambda \sim Beta(\alpha, \alpha)$. This process “softens” the originally binary labels into range $[0, 1]$.  

\begin{figure}[t]
  \centering
  \includegraphics[width=\linewidth]{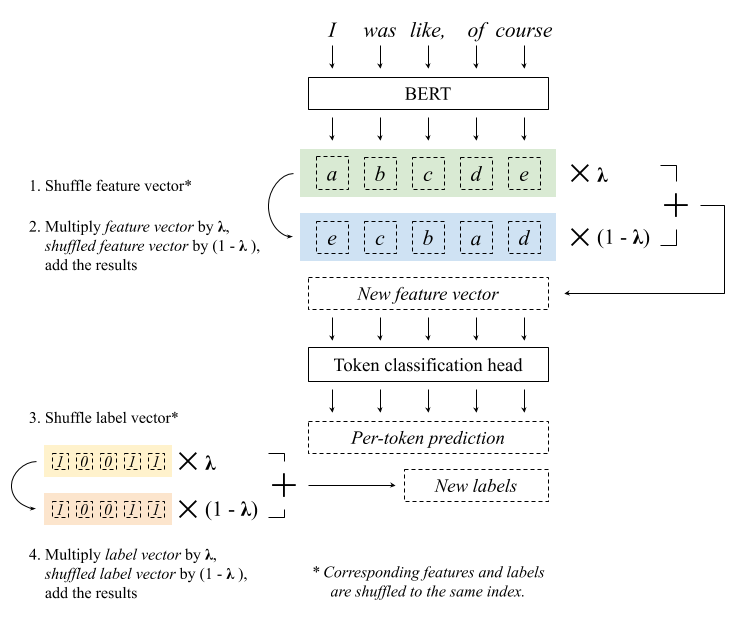}
  \caption{Overview of feature-space regularization for text.}
  \label{fig:architecture}
\end{figure}

For the \textit{mixup} hyperparameter $\alpha$, we empirically chose value a of $0.2$. Figure 3 shows the experimentally observed relationship between $\alpha$ and the normalized counts of the fine-tuned system erroneously tagging disfluent speech as ASR errors.   

$\vec{h_{new}}$ is propagated to the token classification head, whose prediction loss is calculated using binary cross entropy against $\vec{y_{new}}$, instead of the original ground truth label $\vec{y}$.  

\begin{figure}[t]
  \centering
  \includegraphics[width=\linewidth]{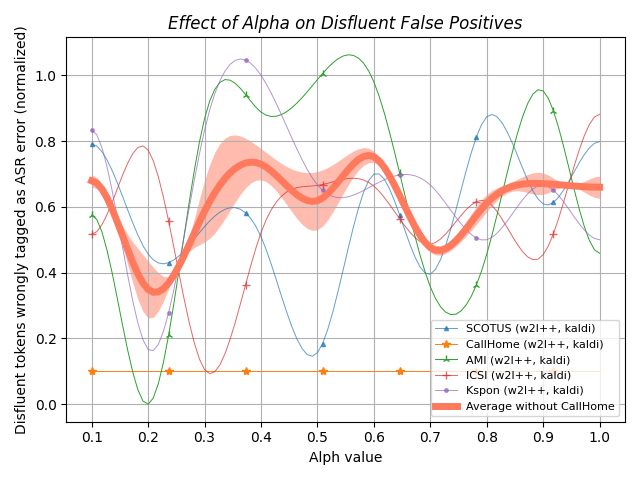}
  \caption{Effect of $\alpha$ and normalized counts at which the system wrongly detects disfluent tokens as ASR errors. Regions within 1 variance of average is shaded.}
  \label{fig:alpha}
\end{figure}

\section{Experiments}

\subsection{Experiment setup}
To obtain ASR outputs on which to perform error detection, we process multiple speech corpora through black box ASR systems. Detailed descriptions of datasets used is provided in the next section. Within the decoded transcript, we tag subword token indices that contain ASR errors to be used as labels during the post-processing fine-tuning. We also tag indices in the transcription with speech disfluencies. To demonstrate the effectiveness of our method across various off-the-shelf ASR environments, we transcribe all available speech corpora with both a traditional ASR system and an end-to-end (E2E) ASR system. For English corpora, we utilize a pretrained LibriSpeech Kaldi \cite{povey_kaldi_2011} model (factorized time-delay neural networks-based chain model) for the traditional system, and a reference wav2letter++ (w2l++) \cite{pratap1812wav2letter++} system for the E2E system. For Korean corpora, we use the same respective architectures trained from scratch on internal data. 

To create error-labeled decoded transcripts, each decoded transcript is compared with corresponding ground truth reference transcription. We use the Wagner-Fischer algorithm \cite{navarro_guided_2001} to build a shortest-distance edit matrix from the decoded transcript to its reference transcription. Traversing the edit matrix from the lowest, rightmost column to the highest, leftmost column yields a sequence of \textit{delete}, \textit{replace}, and \textit{insert} instructions to convert the decoded transcript into the reference transcript. Each character instructed to be deleted and replaced in the decoded transcript is marked as ASR errors.  

Disfluency labels in reference transcripts must be transferred over to corresponding positions within the decoded transcripts. Since the introduction of ASR errors causes the decoder transcripts to be mere approximations of its ground truth transcripts, mapping each character index from the reference transcript to the decoded transcript is not rigorously a defined process. Here we again utilize the Wagner-Fischer algorithm, mapping only the character indices in the decoded transcript without any edit instructions to their corresponding indices in the reference transcript. 

We judge the effectiveness of our regularization method by two criteria: how much it helps in error detection (\textit{Error detection $F_1$)}, and how much it reduces the number of accurately transcribed disfluent speech tagged as errors (\textit{Disfluencies wrongly tagged}).

\subsection{Desciption of datasets}
We tested our proposed methodology on five different speech datasets (4 English, 1 Korean). All speech corpora were provided with reference transcripts hand-annotated with speech disfluency positions.

\subsubsection{SCOTUS}
The SCOTUS corpus \cite{zayats_multi-domain_2014} consists of seven oral arguments between justices and advocates. Nested disfluency tags suggested in \cite{shriberg_preliminaries_1994} were flattened and hand-annotated.  

\subsubsection{CallHome}
The CallHome corpus \cite{linguistic_data_consortium_cabank_2013} is a collection of telephone calls. Each participant in the recording was asked to call a person of their choice. Similar to SCOTUS, nested disfluency tags were flattened and hand-annotated \cite{zayats_multi-domain_2014}.

\subsubsection{AMI Meeting Corpus (AMI)}
The AMI Meeting Corpus \cite{mccowan2005ami} contains recordings of multiple-speaker meetings on a wide range of subjects. Multiple types of crowd-sourced annotations are available, including speech disfluencies, body gestures, and named entities. 

\subsubsection{ICSI Meeting Corpus (ICSI)}
The ICSI Meeting Corpus \cite{janin_icsi_2003} is a collection of multi-person meeting recordings. The ICSI Meeting corpus provides the highest count of disfluency-annotated transcripts out of the five corpora used in this paper.  
\subsubsection{KsponSpeech (Kspon)}
KsponSpeech \cite{bang_ksponspeech_2020} is a collection of spontaneous Korean speech. The corpora is a recording of sub-10 seconds of spontaneous Korean speech. Each recording is labeled with disfluencies and mispronunciations. 

\begin{table}[th]
  \caption{Statistics for each corpus}
  \label{tab:example}
  \centering
  \begin{tabular}{c c c c c}
    \toprule
    \multicolumn{1}{c}{\textbf{Corpus}} & 
    \multicolumn{2}{c}{\textbf{Training samples}} &
    \multicolumn{2}{c}{\textbf{Token level \break error rate}} \\
    \midrule
     & \textit{E2E}   & \textit{Traditional} & \textit{E2E} & \textit{Traditional} \\
    \textbf{SCOTUS}   & $1063$  & $2163$  & $0.12$ & $0.12$ \\
    \textbf{CallHome} & $1420$  & $3058$  & $0.43$ & $0.48$ \\
    \textbf{AMI}      & $4062$  & $6963$  & $0.35$ & $0.45$ \\
    \textbf{ICSI}     & $16389$ & $24165$ & $0.10$ & $0.22$ \\
    \textbf{Kspon}    & $2406$  & $2406$  & $0.06$ & $0.15$ \\
    \bottomrule
  \end{tabular}
  
\end{table}

ASR output for each corpus was randomly split into 8:1:1 \textit{train}:\textit{validation}:\textit{test} splits. Table 1 describes the characteristics of ASR outputs with which we fine-tune BERT. \textit{Training samples} denote the number of sub-word tokenized sequences obtained from each ASR output. \textit{Token level error rate} shows the rate at which subword tokens contain errors in transcription. This latter statistic is purely concerned with processing ASR outputs, and thus conveys a slightly different meaning than conventional ASR error metrics, such as character error rates. Most importantly, words missing in ASR outputs (compared to ground-truth transcriptions) is not tallied in Table 1's token level error rate. This omission was introduced on purpose, because at inference time our model is not expected to introduce new edits but merely tag each subword token as \textit{correct} or \textit{incorrect}.

Number of training examples obtained from a single corpus transcribed with English E2E ASR differs from the number of training examples obtained from the same corpus transcribed with English traditional ASR. This discrepancy stems from the utilized E2E system's tendency to refrain from transcribing words with low confidence. Since the purpose of this study is to demonstrate the effectiveness of our proposed regularization method in various ASR post-processing setups, the discrepancy in the number of training examples from different ASR systems was generated on purpose.

\subsection{Results}
Results of applying our proposed method is presented Table 2. Corpus names in the first column are each marked with the name of ASR system used for transcription. The second and third columns of table 3 show error detection $F_{1}$ scores with and without applying proposed regularization, respectively. The fourth and fifth columns report the number of correctly transcribed disfluencies wrongly tagged as ASR errors.

The proposed regularization in this paper shows greater influence as the number of training data increases (\textit{AMI} and \textit{ICSI}). One surprising outcome is the unchanging scores on the \textit{CallHome} corpus. Analysis of the corpus and high $F_1$ scores in error detection indicate the uniform scores are a result of distinct patterns in ASR errors during transcription; the fine-tuned LM in every setting learns to confidently and consistently pick out only certain tokens as ASR errors.     

\begin{table}[th]
  \caption{All results are subword token level scores, averaged over 6 repeated experiments with different random seeds.}
  \label{tab:example}
  \centering
  \begin{tabular}{c c c | c c}
    \toprule
    \multicolumn{1}{c}{\textbf{Corpus}} &
    \multicolumn{2}{c}{\textbf{Error detection $F_{1}$}} & 
    \multicolumn{2}{c}{\textbf{Disf. wrongly tagged}} \\
    \midrule
    & \textit{Mixup}   & \textit{No mixup} & \textit{Mixup} & \textit{No mixup} \\
    \midrule
    \textit{SCOTUS (w2l++)}   & $\boldsymbol{0.467}$  & $0.462$  & $\boldsymbol{254}$ & $312$ \\
    \textit{SCOTUS (Kaldi)}   & $0.651$  & $\boldsymbol{0.662}$  & $392.5$ & $\boldsymbol{345.75}$ \\
    \textit{CallHome (w2l++)} & $0.910$  & $0.910$  & $164$ & $164$ \\
    \textit{CallHome (Kaldi)} & $0.950$  & $0.950$  & $324$ & $344$ \\
    \textit{AMI (w2l++)}      & $\boldsymbol{0.760}$  & $0.746$  & $\boldsymbol{253.25}$ & $291.25$ \\
    \textit{AMI (Kaldi)}      & $\boldsymbol{0.920}$  & $0.915$  & $\boldsymbol{140}$ & $141.25$ \\
    \textit{ICSI (w2l++)}     & $\boldsymbol{0.347}$ & $0.340$ & $\boldsymbol{1556}$ & $1559.75$ \\
    \textit{ICSI (Kaldi)}     & $\boldsymbol{0.777}$ & $0.771$ & $\boldsymbol{1085}$ & $1124.25$ \\
    \textit{Kspon (w2l++)}    & $\boldsymbol{0.332}$  & $0.308$  & $3$ & $\boldsymbol{2.3}$ \\
    \textit{Kspon (Kaldi)}    & $\boldsymbol{0.465}$  & $0.464$  & $8$ & $\boldsymbol{5}$ \\
    \bottomrule
  \end{tabular}
  
\end{table}

\section{Conclusion}

The discrete nature of textual data has stifled adaptation of \textit{mixup} in natural language processing and speech-to-text domains. In this study we propose a way to effectively apply \textit{mixup} to improve post-processing error detection.

We extensively demonstrate the effect of our proposed approach on various speech corpora. Impact of feature-space \textit{mixup} is especially pronounced in high resource settings. The proposed regularization layer is also computationally inexpensive and can be easily retrofitted to existing systems as a simple module. While we chose the popular BERT LM for our experiments, our method can be applied to fine-tuning any deep LM in a sequence tagging setting. We also highlighted potential uses of resulting LMs in semi-supervised ASR training as pseudo-label filters. Testing the effectiveness of this approach in other sequence-level tasks is left to further research.

\bibliographystyle{IEEEtran}

\bibliography{mybib}


\end{document}